\documentclass{article}

\usepackage[preprint]{neurips_2026}

\usepackage[utf8]{inputenc} 
\usepackage[T1]{fontenc}    
\usepackage{hyperref}       
\usepackage{url}            
\usepackage{booktabs}       
\usepackage{amsfonts}       
\usepackage{nicefrac}       
\usepackage{microtype}      
\usepackage{xcolor}         

\usepackage{microtype}
\usepackage{subcaption}
\usepackage{graphicx}
\usepackage{booktabs} 
\usepackage{tcolorbox}
\usepackage{listings}
\usepackage[T1]{fontenc}  
\usepackage{textcomp}    
\usepackage{lmodern}     
\usepackage{makecell}
\usepackage{xcolor}

\newcommand{\rev}{\textcolor{black}}

\definecolor{backcolour}{rgb}{0.95,0.95,0.96}  
\definecolor{codegreen}{rgb}{0,0.6,0}          
\definecolor{codegray}{rgb}{0.5,0.5,0.5}       
\definecolor{codepurple}{rgb}{0.58,0,0.82}    
\definecolor{codeblue}{rgb}{0.1,0.3,0.7}    
\definecolor{keycolor}{rgb}{0.7, 0.1, 0.1}     
\lstdefinelanguage{YAML}{
  keywords={true,false,null},
  keywordstyle=\color{codepurple}\bfseries,
  ndkeywords={task, events, agents, environment}, 
  ndkeywordstyle=\color{keycolor}\bfseries,
  identifierstyle=\color{black},
  sensitive=false,
  comment=[l]{\#},
  morestring=[b]',
  morestring=[b]"
}

\usepackage{amsmath}
\usepackage{amssymb}
\usepackage{mathtools}
\usepackage{kotex}
\usepackage{multirow}
\usepackage{booktabs}
\usepackage{enumitem}
\usepackage{minted}
\usepackage{threeparttable}

\usepackage{booktabs}   
\usepackage{tabularx}   
\usepackage{array}      

\newcolumntype{L}{>{\raggedright\arraybackslash}m}

\renewcommand{\tabularxcolumn}[1]{m{#1}}

\usepackage[textsize=tiny]{todonotes}

\usepackage[accsupp]{axessibility}  

\usepackage[capitalize,noabbrev]{cleveref}

\usepackage{orcidlink}

\newcommand{\ours}[1]{\textbf{TickingCollab}}
\newcommand{\oursbenchmark}[1]{\textbf{TickingCollabBench}}
\newcommand{\oursagent}[1]{\textbf{TickingCollabAgent}}

\newenvironment{packeditemize}{
\begin{list}{$\bullet$}{
\setlength{\itemsep}{0pt}
\addtolength{\labelwidth}{10pt}
\setlength{\leftmargin}{12pt}
\setlength{\listparindent}{\parindent}
\setlength{\parsep}{2pt}
\setlength{\topsep}{-6pt}}}
{\end{list}
}

\title{Multi-agent Framework for Time-Sensitive Complementary Collaboration in Minecraft}

\author{%
    Juheon Yi, Jinglu Wang, Xiaoyi Zhang, Yan Lu\\
    Microsoft Research Asia \\
    \{jyi,jinglwa,xiaoyizhang,yanlu\}@microsoft.com
}

\begin{document}

\maketitle

\begin{abstract}
We present \oursbenchmark{}, a Minecraft-based multi-agent benchmark for a novel class of \emph{time-sensitive complementary collaboration tasks}. 
Our benchmark reflects four core characteristics of real-world collaboration: agent heterogeneity, mandatory collaboration, dynamic environments, and strict real-time constraints with failure risks. 
To enable this, we develop the \ours{} framework, which supports the generation of diverse dynamic events and abstracts Minecraft's primitive APIs to enable declarative YAML task specifications for composing these events.
Building on this, we design a feasibility-aware automated benchmark generation pipeline, where an LLM drafts structurally diverse task configurations and feasibility verifier filters out invalid ones using approximate constraints.
Evaluations demonstrate that long latency and the inherent difficulty of coordinating under partial observability and agent heterogeneity cause LLMs to frequently fail under dynamic environments and fall significantly short of a global-knowledge oracle.
\end{abstract}

\section{Introduction}
\label{sec:1-intro}

Real-world multi-agent collaboration often requires agents with partial observability to synergistically combine heterogeneous capabilities and complete tasks under strict time constraints. 
For instance, a team of embodied robots with different tools and mobility may need to coordinate to respond to spreading hazards before a rescue deadline.
Similarly, in collaborative work settings, personal agents running on different users' devices may observe only local data and have heterogeneous computing resources, requiring them to jointly process distributed information to respond to users' requests in a timely manner.
However, composing such time-sensitive collaborative scenarios in the real world and evaluating agents at scale is difficult due to safety risks, deployment costs, and limited controllability over environment dynamics.
As a result, many prior works focus on static tasks with shared context, homogeneous agents, and no explicit time-to-failure constraints~\cite{zhuge2024gptswarm, chen2024agentverse, yu2024researchtown}.

To bridge this gap, Minecraft has emerged as a scalable testbed for composing complex tasks and dynamic environments, and systematically controlling agent capabilities such as tools, mobility, and perception.
However, existing Minecraft-based multi-agent collaboration benchmarks~\cite{white2025minecollab, schipper2025pillagerbench, long2024teamcraft, yu2024mineland, dong2024villageragent} still suffer from critical limitations:

\begin{figure*}[t]
    \centering
    \begin{subfigure}[b]{0.32\linewidth}
        \centering
        \includegraphics[width=0.9\linewidth]{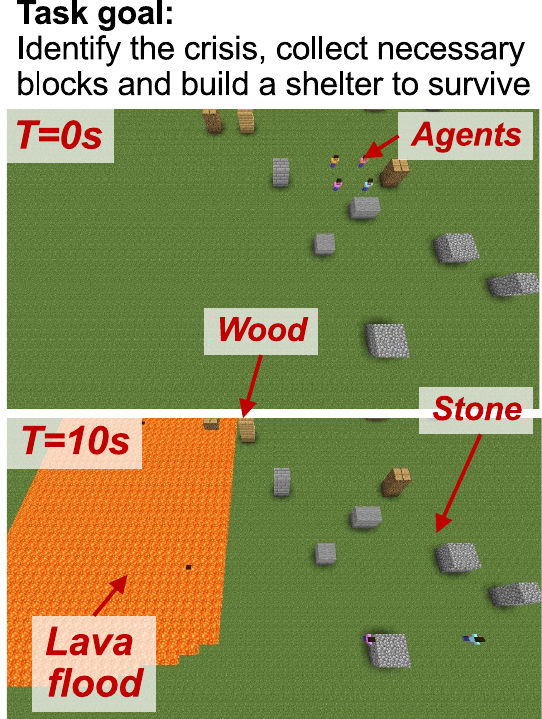}
        \caption{Prepare for a crisis.}
        \label{fig:2-task-overview-a}
    \end{subfigure}
    \hfill
    \begin{subfigure}[b]{0.32\linewidth}
        \centering
        \includegraphics[width=0.9\linewidth]{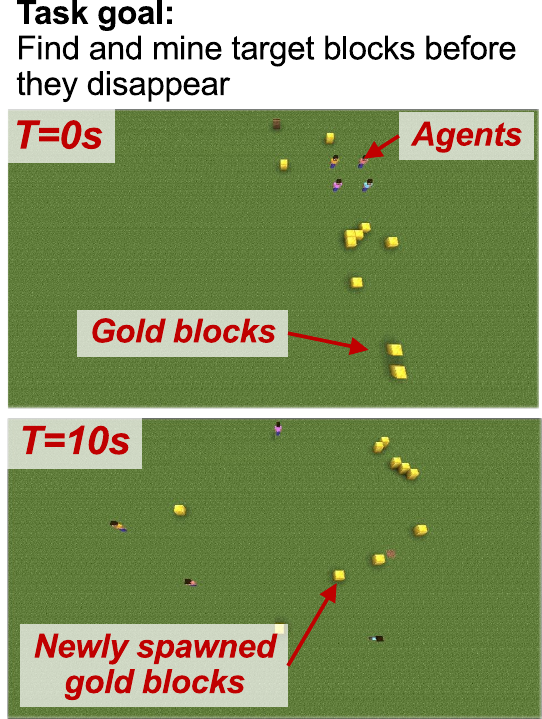}
        \caption{Mine vanishing blocks.}
        \label{fig:2-task-overview-b}
    \end{subfigure}
    \hfill
    \begin{subfigure}[b]{0.32\linewidth}
        \centering
        \includegraphics[width=0.9\linewidth]{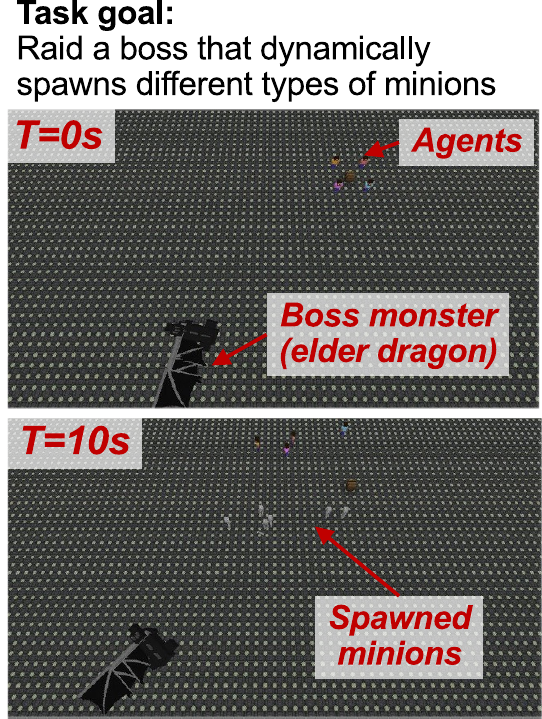}
        \caption{Raid a boss.}
        \label{fig:2-task-overview-c}
    \end{subfigure}
    \hfill
    \begin{subfigure}[b]{1\linewidth}
        \centering
        \includegraphics[width=0.9\linewidth]{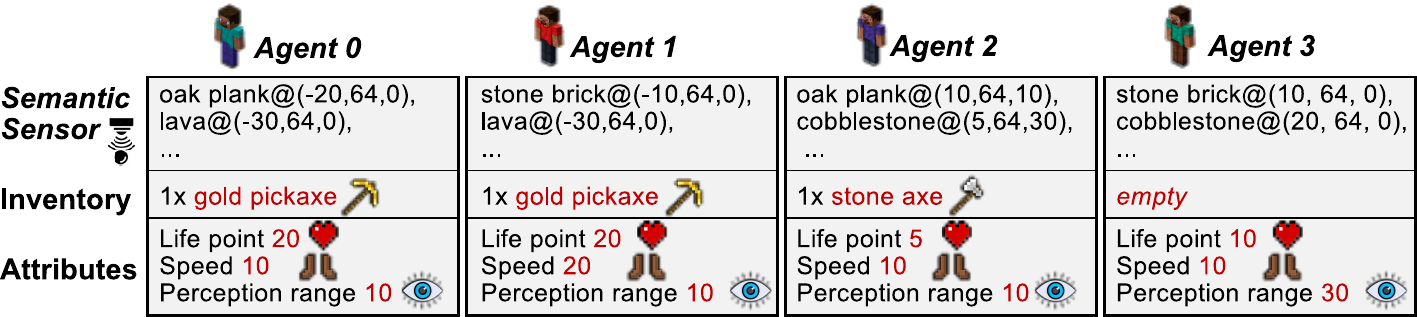}
        \caption{Agent observations and capabilities.}
        \label{fig:2-task-overview-d}
    \end{subfigure}
    \vskip -0.1in
    \caption{Time-sensitive complementary collaboration tasks in \oursbenchmark{}.}
    \label{fig:2-task-overview}
    \vskip -0.3in
\end{figure*}

\begin{packeditemize}
  \item \textbf{Insufficient emphasis on real-time, dynamic collaboration.} 
  Existing tasks often feature homogeneous agents and are individually solvable, rendering genuine collaboration optional rather than mandatory. 
  Furthermore, their static environments and best-effort objectives allow agents to rely on one-time offline planning, facing little to no failure risk from delayed decisions or runtime dynamic events (quantitative analysis in \cref{tb:3-benchmark-stats}).
  
  \item \textbf{Limited framework support for dynamic tasks}.
  Existing Minecraft frameworks are primarily designed for static environments and lack built-in support for runtime dynamic events (e.g., spreading floods, object/monster spawn and despawn). 
  Consequently, to introduce such dynamics, agent developers must utilize low-level Minecraft APIs to build custom server plugins from scratch. 
  This high technical barrier creates severe development overhead, thereby limiting the creation of diverse and complex collaboration scenarios.
\end{packeditemize}

We present \ours{}, a benchmark suite and framework for evaluating LLM agents on a novel class of \emph{time-sensitive complementary collaboration tasks} in Minecraft.
\oursbenchmark{} (\cref{fig:2-task-overview}) targets scenarios where agents with heterogeneous capabilities and partial observability must tightly integrate their complementary skills.
Crucially, the environments continuously change, and failure to rapidly adapt directly causes task failure.
Supported by quantitative comparisons with prior benchmarks (\cref{tb:2-benchmark-comparison,tb:3-benchmark-stats}), we pose a fundamental question:
\emph{can LLMs orchestrate accurate and efficient collaboration across heterogeneous agents, when faced with dynamic environments and real-time failure risks?}

To systematically construct and evaluate time-sensitive complementary collaboration tasks, our \ours{} framework provides three key functionalities:
\begin{packeditemize}
  \item \textbf{Dynamic environment manager.} 
  Developers can declaratively inject complex runtime dynamics (e.g., lava waves, object/monster spawn/despawn) via Minecraft API-free YAML configurations (\cref{list:2-metadata-example}), bypassing the severe overhead of custom plugin development.
  
  \item \textbf{Feasibility-aware automated benchmark generation.} 
  To systematically explore a large and complex parameter space in composing time-sensitive complementary collaboration tasks, we design an automated pipeline where an LLM drafts diverse task configurations and a feasibility verifier filters out invalid ones via approximate constraints.
  
  \item \textbf{Comprehensive evaluation.} The framework isolates LLM's planning accuracy from inference latency via dual execution modes (\emph{synchronous fixed-timestep} vs. \emph{asynchronous real-time}), while supporting parallel simulation and fine-grained system cost logging.
\end{packeditemize}

We evaluate our benchmark using a baseline multi-agent collaboration scheme (\oursagent{}) with two distinct coordination policies (centralized and distributed) motivated from prior works~\cite{long2024teamcraft, white2025minecollab}.
Our evaluation reveals that LLM inference latency poses a critical bottleneck in real-time asynchronous execution, frequently causing task failures due to time constraint violations. 
Furthermore, while centralized coordination outperforms distributed topologies by mitigating communication and inference overheads, it still underperforms an oracle--a non-LLM solution that leverages global, ground-truth access to the dynamic environment and human-crafted scheduling rules.
These findings underscore the challenges of heterogeneous multi-agent planning under partial observability and necessitate efficient LLM inference and multi-agent coordination policies in dynamic environments.
\section{\oursbenchmark{} Benchmark Suite}
\label{sec:2-benchmark}

\begin{table*}[t]
\caption{Comparison of prior Minecraft agent benchmarks and \oursbenchmark{}. \\$\triangle$: partially covered for the subset of tasks listed in parentheses.}
\label{tb:2-benchmark-comparison}
\vskip -0.08in
\centering
\small
\resizebox{1\textwidth}{!}{
    \begin{threeparttable}
        \begin{tabular}{clcccc}
        \toprule
         & \textbf{Benchmark} & \textbf{\shortstack{Heterogeneous\\agent capabilities?}} & \textbf{\shortstack{Mandatory\\collaboration?}} & \textbf{\shortstack{Dynamic\\environment?}} & \textbf{\shortstack{Real-time constraints\\(or failure risks)?}} \\
        \midrule
        
        \multirow{4}{*}{\shortstack[l]{Single\\agent}}
         & MineRL~\cite{guss2019minerl} & $\times$ & $\times$ & $\times$ & $\times$ \\
         & MineDojo~\cite{fan2022minedojo} & $\times$ & $\times$ & $\triangle$ (combat) & $\triangle$ (combat, survive) \\
         & Odyssey~\cite{liu2024odyssey} & $\times$ & $\times$ & $\triangle$ (combat) & $\triangle$ (combat, survive)      \\
         & MCU~\cite{zheng2025mcu}     & $\times$ & $\times$ & $\triangle$ (combat) & $\triangle$ (combat, survive) \\
        \midrule
        
        \multirow{6}{*}{\shortstack[l]{Multi-\\agent}}
         & MineLand~\cite{yu2024mineland} & $\times$ & $\times$ & $\triangle$ (combat) & $\triangle$ (combat, survive) \\
         & TeamCraft~\cite{long2024teamcraft} & $\bigcirc$ (different items) & $\bigcirc$ & $\times$ & $\times$ \\
         & MineCollab~\cite{white2025minecollab}  & $\bigcirc$ (different items) & $\triangle$ (cook, craft) & $\times$  & $\times$      \\
         & PillagerBench~\cite{schipper2025pillagerbench} & $\times$ & $\times$ & $\bigcirc$ & $\bigcirc$ (vs. opponents) \\
         & VillagerBench~\cite{dong2024villageragent}       & $\times$ & $\triangle$ (escape room) & $\triangle$ (harvest) & $\times$ \\
         \cmidrule{2-6}
         & {\bf \oursbenchmark{}} & $\bigcirc$ & $\bigcirc$ & $\bigcirc$ & $\bigcirc$ \\ 
        \bottomrule
        \end{tabular}
    \end{threeparttable}
}
\vskip -0.1in
\end{table*}

\oursbenchmark{} consists of three representative \emph{time-sensitive complementary collaboration tasks} (\cref{fig:2-task-overview}), where agents with heterogeneous capabilities must coordinate to achieve a global objective under partial observability of the dynamic environment.
These tasks are designed as controlled analogues of the real-world collaboration scenarios discussed in \cref{sec:1-intro}, emphasizing four key properties that are underrepresented in previous Minecraft multi-agent benchmarks (see \cref{tb:2-benchmark-comparison} for a detailed comparison):

\begin{packeditemize}
    \item \textbf{Heterogeneous capabilities.} 
    Agents differ in action-relevant attributes and resources (e.g., perception range, speed, health, tools), necessitating complex complementary roles that cannot be captured by simple inventory differences featured in prior benchmarks.
    
    \item \textbf{Mandatory collaboration.} 
    Tasks are constructed so that success requires coordinating complementary capabilities, rather than simply scaling identical agents.
    
    \item \textbf{Dynamic environments.} 
    Environments continuously change at runtime, invalidating one-shot offline plans and requiring online adaptation.
    
    \item \textbf{Real-time constraints.} 
    Unlike prior penalty-free settings, delayed decision-making directly causes task failure, demanding timely execution.
\end{packeditemize}

\ours{} framework provides a declarative interface for specifying these tasks. 
This enables an automated benchmark generation pipeline: an LLM first generates diverse configurations incorporating the four key properties, which are then filtered based on feasibility criteria to yield 634 valid tasks for \oursbenchmark{} (details in \cref{subsec:3-metadata}).

\subsection{Task Suite}
\label{subsec:2-tasks}

\noindent {\bf Task \#1: Prepare for a crisis (\cref{fig:2-task-overview-a})}.
Agents must identify an approaching crisis (e.g., lava flood, avalanche) and collaboratively gather appropriate materials scattered in the map to build an appropriate survival shelter (e.g., stone instead of flammable wood) before crisis impact. 
Agents possess different mining tools (e.g., axes for woods, pickaxes for metal ores), perception ranges, and movement speeds. 
Survival depends on efficient role allocation, such as utilizing long-perception agents as scouts while faster agents collect distant blocks. 
Unlike prior \emph{construction tasks} that ignore time constraints and failure risks, this task demands timely execution and efficient coordination of heterogeneous agents.

\noindent {\bf Task \#2: Mine vanishing blocks (\cref{fig:2-task-overview-b})}.
Agents must mine target quotas for multiple block types that randomly appear and vanish after type-specific lifetimes.
Given distinct movement speeds, perception ranges, and heterogeneous mining tools that dictate both block compatibility and mining efficiency (e.g., wood requires an axe, gold ore requires at least an iron-tier pickaxe, and higher-tier tools like diamond yield faster mining rates), agents must optimally assign targets by calculating travel and mining times against block lifetimes to avoid wasted effort. 
While previous \emph{harvesting tasks} mostly feature static block placements and uniform agents, our task requires dynamic, capability-aware assignment.

\noindent {\bf Task \#3: Raid a boss (\cref{fig:2-task-overview-c})}.
Agents must defeat a boss monster that dynamically spawns various minions with different life points (HP) and damages. 
Agents differ in base life points and weapons with type-specific damage multipliers. 
Agents must jointly optimize target assignments based on type advantages and survivability, while strategically disengaging to consume health potions at scattered chests. 
Whereas prior \emph{combat tasks} typically assume statically generated monsters and homogeneous agents, ours introduces dynamic enemy spawns and demands intricate combat coordination among diverse roles.

\subsection{Collaboration Difficulty Metrics}
\label{subsec:2-metrics}

\begin{table}[t]
\centering
\caption{Comparison of multi-agent benchmark statistics. 
$\uparrow$ and $\downarrow$ indicate whether lower or higher values imply a more challenging task, respectively.
}
\vskip 0.02in
\label{tb:3-benchmark-stats}
\resizebox{\columnwidth}{!}{%
\begin{tabular}{l|ccc|cccc|ccc|ccc}
\toprule
& \multicolumn{3}{c|}{MineLand~\cite{yu2024mineland}} 
& \multicolumn{4}{c|}{TeamCraft~\cite{long2024teamcraft}} 
& \multicolumn{3}{c|}{MineCollab~\cite{white2025minecollab}} 
& \multicolumn{3}{c}{\textbf{\oursbenchmark{}}} \\
\midrule

& Combat & Harvest & Build
& Break & Build & Farm & Smelt
& Build & Cook & Craft
& Prepare & Mine & Raid \\
\midrule
$\mathcal{H}\uparrow$ 
& 0 & 0 & 0
& 0.72 & 0.43 & 0.39 & 0.68
& 0 & 0.43 & 0.44
& 0.78 & 0.72 & 0.31 \\

$\mathcal{N}\uparrow$
& 0 & 0 & 0
& 0 & 0.54 & 1.13 & 0
& 0 & 1.87 & 1.76
& 1.11 & 1.42 & 1.39 \\

$\mathcal{D}\uparrow$
& 0 & 0 & 0
& 0 & 0 & 0 & 0
& 0 & 0 & 0
& 1.79 & 2.98 & 0.38 \\

$\tau\downarrow$
& 11.51s$^*$ & $\infty$ & $\infty$
& $\infty$ & $\infty$ & $\infty$ & $\infty$
& $\infty$ & $\infty$ & $\infty$
& 44.03s & 33.31s & 25.37s \\

\bottomrule
\end{tabular}%
}
{\scriptsize \raggedright * Samples with $\tau = \infty$ are excluded. \par}
\vskip -0.2in
\end{table}

We define four collaboration difficulty metrics to quantitatively evaluate how \oursbenchmark{} reflects time-sensitive, complementary collaboration.

\begin{packeditemize}
    \setlength{\abovedisplayskip}{1pt}
    \setlength{\belowdisplayskip}{1pt}
    \setlength{\abovedisplayshortskip}{0pt}
    \setlength{\belowdisplayshortskip}{0pt}

    \item \textbf{Agent heterogeneity ($\mathcal{H}$)} 
    measures the average pairwise normalized distance of attributes across all unique agent pairs $P$:
    \begin{equation*}
        \resizebox{0.45\linewidth}{!}{$
            \mathcal{H} =
            \frac{1}{|P|}
            \sum_{(i, j) \in P}
            \left(
                \frac{1}{|K|}
                \sum_{k \in K}
                \delta_k(a_i, a_j)
            \right).
        $}
    \end{equation*}
    For each attribute $k \in K$, the distance $\delta_k$ handles continuous values (e.g., HP) via min-max normalization and sets (e.g., inventory items) via Jaccard distance:
    \begin{equation*}
        \resizebox{0.5\linewidth}{!}{$
        \delta_k(a_i, a_j) = 
        \begin{cases} 
            \frac{|v_{ik} - v_{jk}|}{v_k^{max} - v_k^{min}} & \text{if } k \text{ is continuous} \\
            1 - \frac{|S_{ik} \cap S_{jk}|}{|S_{ik} \cup S_{jk}|} & \text{if } k \text{ is a set }
        \end{cases}
        $}
    \end{equation*}
    where $v_k^{max}$ and $v_k^{min}$ denote predefined bounds of the parameter space. Thus, $\mathcal{H} \in [0, 1]$, where $1$ implies maximal distinctiveness.

    \item \textbf{Collaboration necessity ($\mathcal{N}$)} 
    estimates the ratio of the total task workload to the maximum single-agent capacity:
    \begin{equation*}
        \resizebox{0.45\linewidth}{!}{$
        \mathcal{N} = \min_{a \in A} \left( \sum_{k} \frac{\text{Workload}_k}{\text{Throughput}_{a,k}} \right) / T_{max}.
        $}
    \end{equation*}
    The inner sum computes the time required for the most capable single agent $a \in A$ to sequentially process all targets $k$. Here, $\text{Workload}_k$ is the required block count (Tasks \#1, \#2) or total enemy HP (Task \#3), while $\text{Throughput}_{a,k}$ is agent $a$'s mining speed or damage-per-second (DPS). 
    Note that $\mathcal{N}$ is a conservative lower bound, as the time required for agent planning and movement are omitted; thus, $\mathcal{N}>1$ strictly guarantees that collaboration is mandatory.

    \item \textbf{Environment dynamicity ($\mathcal{D}$)} is quantified as:
    \begin{equation*}
        \resizebox{0.5\linewidth}{!}{$
        \mathcal{D} = \left(\text{Total Environment State Changes}\right)/T_{max}. 
         $}
    \end{equation*}
    A ``state change'' is any environment modification independent of agent actions, such as entity spawns/despawns (Tasks \#2, \#3) or crisis spread (Task \#1). 
    Higher $\mathcal{D}$ necessitates continuous online replanning.

    \item \textbf{Time-to-failure ($\tau$)} 
    measures the time window before irreversible failure occurs, which is defined as: crisis arrival time (Task \#1), minimum block lifespan (Task \#2), or the time for spawned enemies to defeat all agents, calculated as $\frac{\text{Total Agent HP}}{\text{Total Enemy DPS}}$ averaged across spawn events (Task \#3). Lower $\tau$ demands faster decision-making.
\end{packeditemize}

\begin{figure}[t]
\centering
\begin{subfigure}[b]{0.24\textwidth}
    \includegraphics[width=\linewidth]{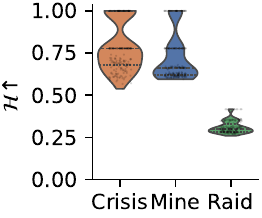}
    \caption{Heterogeneity ($\mathcal{H}\!\uparrow$).}
\end{subfigure}%
\begin{subfigure}[b]{0.24\textwidth}
    \includegraphics[width=\linewidth]{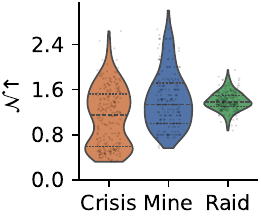}
    \caption{Necessity ($\mathcal{N}\!\uparrow$).}
\end{subfigure}%
\begin{subfigure}[b]{0.24\textwidth}
    \includegraphics[width=\linewidth]{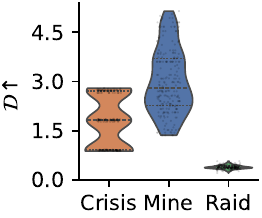}
    \caption{Dynamicity ($\mathcal{D}\!\uparrow$).}
\end{subfigure}%
\begin{subfigure}[b]{0.24\textwidth}
    \includegraphics[width=\linewidth]{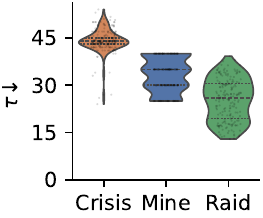}
    \caption{Time-to-failure ($\tau\!\downarrow$).}
\end{subfigure}
\vspace{-5pt}
\caption{
Distribution of collaboration difficulty metrics in \oursbenchmark{}. Internal lines show quartiles (Q1, median, Q3); overlaid dots show individual configurations.
}
\label{fig:2-metric_distributions}
\vskip -0.15in
\end{figure}

As shown in \cref{tb:3-benchmark-stats}, prior Minecraft benchmarks typically feature static environments, negligible failure risks, and limited agent heterogeneity (differing only in inventories), rarely necessitating true collaboration outside a few exceptions (e.g., crafting or cooking in MineCollab~\cite{white2025minecollab}, combat in MineLand~\cite{yu2024mineland}). 
In contrast, \oursbenchmark{} yields significantly higher metric scores, indicating far more challenging collaboration tasks.
\cref{fig:2-metric_distributions} and \cref{tb:2-param_diversity} further show that the generated configurations cover broad ranges of difficulty and task parameters.
Across the three tasks, these distributions reflect diverse collaboration challenges that vary in their focus (e.g., strict time-to-failure constraints in ``Raid a boss'' task, and agent heterogeneity in the other two).

\begin{table}[t]
\centering
\caption{Parameter distributions (min\,/\,mean${\pm}$std\,/\,max for continuous values).}
\label{tb:2-param_diversity}
\resizebox{\columnwidth}{!}{%
\begin{tabular}{@{}lp{3cm}@{\;\;\vrule\;\;}lp{3.8cm}@{\;\;\vrule\;\;}lp{3.5cm}@{}}
\toprule
\multicolumn{2}{@{}l}{\textbf{Prepare for a crisis}} & \multicolumn{2}{l}{\textbf{Mine vanishing blocks}} & \multicolumn{2}{l@{}}{\textbf{Raid a boss}} \\
\midrule
\# Agents & 2 / $3.88{\pm}1.90$ / 8 & \# Agents & 2 / $5.12{\pm}2.25$ / 8 & \# Agents & 3 / $5.93{\pm}1.54$ / 8 \\
Crisis type & lava 30\%, snow 32\%, water 38\% & Tool tier & \begin{tabular}[t]{@{}l@{}}golden 14\%, iron 31\%,\\ stone 32\%, wooden 23\%\end{tabular} & Boss HP & \begin{tabular}[t]{@{}l@{}}210 / $241.53{\pm}18.73$\\ / 280\end{tabular}  \\
Crisis speed & 1 / $1.98{\pm}0.82$ / 3 & \# Target types & 2: 33\%, 3: 36\%, 4: 32\% & Minion HP & 25 / $34.60{\pm}3.87$ / 40 \\
Agent speed & 3 / $4.49{\pm}1.09$ / 6 & Block lifetime & 25 / $33.30{\pm}5.36$ / 40 & \# Minions/wave & 2 / $2.50{\pm}0.56$ / 4 \\
 &  & \# Blocks/wave & 8 / $9.03{\pm}0.70$ / 10 & Wave interval & 8 / $11.34{\pm}1.88$ / 16 \\
\bottomrule
\end{tabular}}
\end{table}
\section{\ours{} Framework}
\label{sec:3-framework}

\begin{figure}[ht]
\begin{center}
\centerline{\includegraphics[width=0.9\columnwidth]{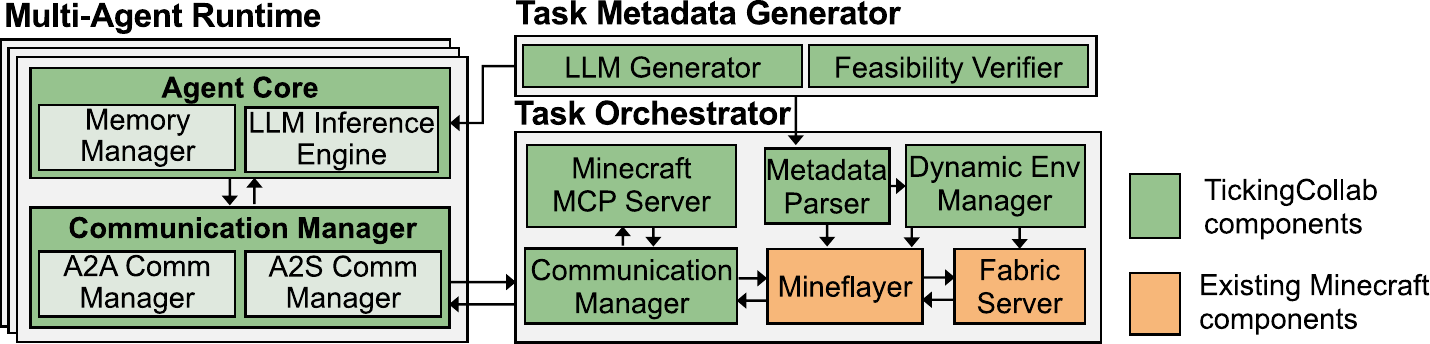}}
\caption{
\ours{} framework architecture. 
}
\label{fig:3-framework-architecture}
\end{center}
\vskip -0.25in
\end{figure}

\cref{fig:3-framework-architecture} illustrates the \ours{} framework architecture. 

\begin{packeditemize}
    \setlength{\abovedisplayskip}{3pt}
    \setlength{\belowdisplayskip}{3pt}
    \setlength{\abovedisplayshortskip}{0pt}
    \setlength{\belowdisplayshortskip}{0pt}

    \item \textbf{Task metadata generator} (\cref{subsec:3-metadata}) 
    provides descriptive task composition via declarative YAML without requiring Minecraft API expertise, enabling an automated generation pipeline to construct \oursbenchmark{}.

    \item \textbf{Task orchestrator} (\cref{subsec:3-server}) 
    translates the metadata into dynamic runtime events via the \emph{dynamic environment manager} and bridges agents to the Minecraft server using Mineflayer~\cite{mineflayer}. It supports both synchronous (fixed-timestep) and asynchronous (real-time) execution to decouple the LLM agent's reasoning accuracy from latency.

    \item \textbf{Multi-agent runtime} (\cref{subsec:3-agent}) 
    provides a modular abstraction for the \emph{agent core} and \emph{communication manager}, streamlining the development of custom collaborative agents.
\end{packeditemize}

\subsection{Task Metadata Generator and Automated Benchmark Generation}
\label{subsec:3-metadata}

\begin{lstlisting}[caption={Example metadata (``prepare for a crisis'', full list in Supp.~\cref{sec:suppl-metadata-schema}).\vspace{-3pt}},  frame=single, breaklines=true,  float=h, label={list:2-metadata-example},
aboveskip=0pt, belowskip=-15pt] %-15, -10

task:
  - goal: "Identify the origin and type of the crisis, and gather necessary blocks to build a shelter for survival."
environment:
    -{type: cobblestone, position: [-5,64,15], num_blocks: 10},
    - {type: oak_log, position: [-8,64,10], num_blocks: 8},
    ...
events: 
    {id: lava_wave, trigger: {start: 5, end: 100}, actions: {type: progressive_fill, block: lava, area: {min: [-40,64,-20], max: [40,65,20]}, direction: east, speed: 2}
agents:  
    {name: MineflayerBot0, position: [8,64,18], inventory: {gold_pickaxe: 1}, capabilities: {perception range: 10; speed: 10}, 
    ... 
\end{lstlisting}



\cref{list:2-metadata-example} illustrates the task metadata structure, comprising four key fields: 
(i) \emph{task description}, 
(ii) \emph{environment layout}, 
(iii) \emph{events} (with dynamic triggers and patterns), and 
(iv) \emph{agents}.
While our declarative interface abstracts away complex Minecraft API programming, manual task design still remains challenging due to the massive parameter space (e.g., jointly aligning crisis type and speed, necessary survival block placements, and agents' speed, perception range, and mining tools).
Prior Minecraft multi-agent benchmarks typically perform naive parameter sweeping (e.g., merely adjusting $N$ identical agents vs. $M$ skeletons) under static environments, yielding trivial variations that fail to meaningfully emphasize collaboration.

To overcome this challenge, we design a \emph{feasibility-aware automated generation pipeline} to construct \oursbenchmark{} at scale. 
First, users define task goals and parameter spaces via a task metadata template (example in Supp. \cref{sec:suppl-user-specification}). 
Using this, the LLM navigates the parameter space to draft structurally diverse configurations spanning various environments, agent compositions, and difficulty levels. 
The LLM may occasionally generate unsolvable task configurations (e.g., requiring gold ore mining without pickaxes, or setting enemy HP beyond the agents' maximum damage output); thus, a \emph{feasibility verifier} screens out invalid configurations using constraints that model the task feasibility.

\cref{tb:2-generation} details the task parameter space (designed to reflect the four properties in \cref{tb:2-benchmark-comparison}) and the feasibility verification criteria. 
Even setting aside runtime stochasticity (e.g., random spawn positions and agent trajectories), optimal target assignment and sequencing (e.g., which blocks to mine) remain intractable.
With heterogeneous agent capabilities and strict time constraints, this reduces to an NP-hard routing and allocation problem (analogous to the multi-agent traveling salesman problem), precluding exact guarantees.
We thus introduce margins $\alpha, \beta, \gamma$ to define approximate feasibility constraints that also control task difficulty.
Generating 250 configurations per task with GPT-5.1 and filtering them with margin values 2.0 yielded 634 valid configurations (225, 219, and 190 for Tasks \#1, \#2, and \#3); see Supp.~\cref{sec:suppl-margin_sensitivity} for margin sensitivity analysis and \cref{sec:suppl-dataset-ex} for generated examples.

\begin{figure}[!htbp] 
\centering
\begin{minipage}{\columnwidth}
    \centering
    \captionsetup{type=table}
    \caption{\ours{}'s automated benchmark generation: parameter space and feasibility verification criteria. Details on each variable in verification criteria are in Supp.-\cref{sec:supp-verification}.}
    \label{tb:2-generation}
    \vskip -0.05in
    \centering
    \footnotesize

    \renewcommand{\tabularxcolumn}[1]{m{#1}}
    
    \newcolumntype{Y}{>{\hsize=1.3\hsize\linewidth=\hsize\raggedright\arraybackslash}X}
    \newcolumntype{Z}{>{\hsize=0.7\hsize\linewidth=\hsize\raggedright\arraybackslash}X}
    
    \newlist{tabitem}{itemize}{1}
    \setlist[tabitem]{nosep, leftmargin=*, label=-, after=\strut}
    
    \begin{tabularx}{1\textwidth}{Y Z}
    \toprule
    \textbf{Parameter space} & \textbf{Verification criteria} \\
    \midrule
    \multicolumn{2}{l}{\underline{\textit{\textbf{Task \#1: Prepare for a crisis}}}} \\
    \noalign{\vskip 2pt}
    \begin{tabitem}
        \item \textbf{Crisis config:} $\mathcal{C} = (Type, P, V, H)$ -- $Type$: crisis type (e.g., lava, water, snow); $P$, $V$, $H$: origin, speed, height.
        \item \textbf{Agent config:} $\mathcal{A}_n = (V_n, R_n, E_n)$ -- $V_n, R_n$: movement speed and perception range; $E_n$: mining tool with tier $\tau_{E_n}$. Block type $i$ is mineable only if $\tau_{E_n} \geq \tau_{req}(i)$, and takes time $T_{mine}^{n,i}$.
        \item \textbf{Block placements:} $\mathcal{I} = \{ (Type_j, Loc_j) \}$ -- types and locations.
        \item \textbf{Simulation duration:} $T_{max}$ steps.
    \end{tabitem} 
    & 
    \begin{tabitem}
        \vspace{-5pt}
        \item \textbf{Necessary mining items?}
        $ \max_{n}(\tau_{E_n}) \ge \tau_{req}(b), \; \forall b \in \mathcal{B}_{target}$ 
        \item \textbf{Sufficient survival blocks?}
        $\textstyle N_{required} \geq (H+1) \cdot N_{agent}$ 
        \item \textbf{Sufficient preparation time?}
        $ T_{crisis} \geq \alpha \cdot (T_{gather}+T_{construct})$ \; ($\alpha$: time margin)
    \end{tabitem} 
    \\
    \midrule
    \multicolumn{2}{l}{\underline{\textit{\textbf{Task \#2: Mine vanishing blocks}}}} \\
    \noalign{\vskip 2pt}
    \begin{tabitem}
        \item \textbf{Target blocks:} $\mathcal{B}_i = (Type_i, N_{goal}^{i})$ -- type and count.
        \item \textbf{Agent config:} $\mathcal{A}_n = (V_n, R_n, E_n)$ -- same as Task \#1.
        \item \textbf{Spawn pattern:} $\mathcal{S}_i = (T_{start}^i, T_{end}^i, T_{int}^i, T_{life}^i)$ -- start, end, interval, lifetime.
        \item \textbf{Simulation duration:} $T_{max}$ steps.
    \end{tabitem}
    & 
    \begin{tabitem}
        \vspace{-10pt}
        \item \textbf{Necessary mining items?}
        $ \max_{n}(\tau_{E_n}) \ge \tau_{req}(b), \; \forall b \in \mathcal{B}_{target}$ 
        \item \textbf{Sufficient block spawns?}
        $\textstyle N_{spawn}^i \geq N_{goal}^i,\;\forall i$
        \item \textbf{Sufficient block lifetimes?}
        $\beta \cdot (\min_{n} (T_{move}+T_{mine}^{n,i})) \leq T_{life}^i, \; \forall i$ \; ($\beta$: time margin)
    \end{tabitem} \\
    \midrule
    \multicolumn{2}{l}{\underline{\textit{\textbf{Task \#3: Raid a boss}}}} \\
    \noalign{\vskip 2pt}
    \begin{tabitem}
        \item \textbf{Boss config:} $\mathcal{B} = (HP_{B}, D_{B})$ -- boss life point and damage.
        \item \textbf{Minion spawns:} $\mathcal{P}_i = (T_{start}^{i}, T_{end}^{i}, T_{int}^{i}, N^{i}, HP_{M}^{i}, D_{M}^{i})$ -- start, end, interval of $i$-th minion spawn events; $N^{i}$: minion count per spawn; $HP_{M}^{i}$, $D_{M}^{i}$: minion life point, damage.
        \item \textbf{Agent config:} $\mathcal{A}_n = (HP_n, D_n)$ -- life point and damage.
        \item \textbf{Simulation duration:} $T_{max}$ steps.
    \end{tabitem} 
    & 
    \begin{tabitem}
        \item \textbf{Sufficient agent damage?}
        $ \gamma \cdot T_{max} \cdot \sum_{n} D_n \geq HP_{B} + \sum_i ( N_i\cdot HP_M^i)$ \; ($\gamma$: efficiency margin)
    \end{tabitem} 
    \\
    \bottomrule
    \end{tabularx}

\end{minipage}

\end{figure}

\subsection{Task Orchestrator}
\label{subsec:3-server}


\noindent {\bf Component \#1: Dynamic environment manager.}
We build a plugin abstracting Minecraft’s complex primitive APIs (e.g., \texttt{/effect}, \texttt{/attribute}) to automatically compose and manage dynamic environment specified in metadata.
Unlike prior benchmarks reliant on static environments, this manager handles diverse, complex runtime dynamics.
It is also highly extensible, allowing developers to inject custom interaction mechanisms not natively supported by Minecraft (e.g., an inventory weight system, region-specific debuffs).

\noindent {\bf Component \#2: Communication manager.}
This module bridges the agent runtime and Mineflayer-controlled bots~\cite{mineflayer} connected to Minecraft server.
It maps high-level agent actions to predefined JavaScript code executed on the Fabric server (Supp.~\cref{sec:suppl-agent-action}) and returns structured observations (agent status and sensor data) for the next decision step.

\noindent {\bf Component \#3: Multi-agent collaboration evaluator.}
To enable in-depth analysis of the agent's reasoning capabilities and system-level costs (\cref{sec:4-experiments}), our framework supports parallel simulation execution, comprehensive logging (e.g., number of inferences and inference latency, token usage, communication overhead), and two distinct evaluation modes:

\begin{packeditemize}
    \item {\bf Synchronous (fixed-timestep) mode} pauses the simulation during LLM inference. It isolates and evaluates pure decision-making accuracy by ignoring inference latency.
    
    \item {\bf Asynchronous (real-time) mode} runs the simulation continuously. It evaluates both accuracy and latency, as prolonged LLM inference causes actions to become stale against the rapidly changing environment, directly impacting task success.
\end{packeditemize}

\subsection{Multi-agent Runtime and \oursagent{}}
\label{subsec:3-agent}

Our modular runtime decouples the agent core (LLM/memory) from communication modules, enabling flexible logic customization.
We implement a baseline \oursagent{} with two distinct coordination policies (system prompts and toolsets in Supp. \cref{sec:suppl-system-prompts,sec:suppl-agent-action}):

\begin{packeditemize}
    \item \textbf{\oursagent{}-centralized} (motivated by TeamCraft~\cite{long2024teamcraft}).
    A designated \emph{master} agent aggregates all peer agents' observations, performs joint planning, and dispatches actions. While centralizing global observations maximizes joint planning quality, it introduces synchronization bottlenecks; the master must wait for all agents to report back before replanning, leaving faster agents idle due to varying action latencies.

    \item \textbf{\oursagent{}-distributed} (motivated by MineCollab~\cite{white2025minecollab}).
    Agents plan independently in parallel, resolving shared decisions (e.g., role allocation) via a propose--wait--act negotiation protocol. 
    We relax MineCollab's 1-to-1 messaging restriction to support selective multi-agent broadcasting, facilitating efficient coordination under time constraints.
\end{packeditemize}

\section{Experiments}
\label{sec:4-experiments}

We host two LLMs (GPT-5.1 and DeepSeek-R1~\cite{guo2025deepseek}) in Azure AI Foundry as the backbone of \oursagent{}, and evaluate their task success rates on our benchmark.
We use an Ubuntu 22.04.5 machine with an AMD EPYC 9V84 CPU (80 logical CPUs), 629 GiB RAM.
\rev{
We implement our \ours{} framework on top of Minecraft Java Edition 1.19, using a Fabric server (loader v0.14.18) and Mineflayer 4.14.0 (unofficial third-party modifications to Minecraft) as the bot control interface.
}

\begin{figure}[ht]
\centering
\begin{minipage}[t]{0.9\linewidth}
    \vspace{-5pt}
    \centering
    \captionof{table}{Average task success rate of \oursagent{}.}
    \label{tb:4-success-rate}
    \vspace{-5pt}
    \setlength{\tabcolsep}{6pt}
    \resizebox{\linewidth}{!}{%
    \begin{tabular}{l cc cc cc cc c}
    \toprule
    & \multicolumn{4}{c}{GPT-5.1}
    & \multicolumn{4}{c}{DeepSeek-R1}
    &  \\
    \cmidrule(lr){2-5} \cmidrule(lr){6-9}
    & \multicolumn{2}{c}{Centralized}
    & \multicolumn{2}{c}{Distributed}
    & \multicolumn{2}{c}{Centralized}
    & \multicolumn{2}{c}{Distributed}
    & \textbf{Oracle} \\
    \cmidrule(lr){2-3} \cmidrule(lr){4-5}
    \cmidrule(lr){6-7} \cmidrule(lr){8-9}
    Task
    & Sync & Async
    & Sync & Async
    & Sync & Async
    & Sync & Async
    & \\
    \midrule
    \makecell[l]{Prepare for a crisis}
    & 0.42 & 0.15
    & 0.24 & 0.02
    & 0.26 & 0.02
    & 0.03 & 0.01
    & 0.91 \\
    \midrule
    \makecell[l]{Mine vanishing blocks}
    & 0.62 & 0.05
    & 0.54 & 0.00
    & 0.65 & 0.00
    & 0.35 & 0.01
    & 0.80 \\
    \midrule
    \makecell[l]{Raid a boss}
    & 0.28 & 0.06
    & 0.40 & 0.03
    & 0.37 & 0.01
    & 0.29 & 0.02
    & 0.59 \\
    \bottomrule
    \end{tabular}%
    } 
\end{minipage}
\vskip -0.1in
\end{figure}


\subsection{Overall Task Success Rate}

\cref{tb:4-success-rate} shows the overall task success rates across various  tasks, LLMs, and execution modes. 
To evaluate the LLM's planning accuracy, we designed an oracle solution where a centralized agent has access to ground-truth task metadata and runtime events:
\begin{packeditemize}
    \item \textbf{Task \#1 Prepare for a crisis}: the oracle retrieves the crisis type, origin, and speed from the task metadata at the beginning of the simulation.
    \item \textbf{Task \#2 Mine vanishing blocks \& Task \#3 Raid a boss}: the oracle receives real-time object spawn/despawn events from task orchestrator (\cref{subsec:3-server}) during runtime.
\end{packeditemize}
Based on this information, the oracle plans the actions of individual agents. Since scheduling heterogeneous agents is an NP-hard problem (analogous to the multi-agent traveling salesman problem), we designed an efficient, non-LLM-based task allocation algorithm using handcrafted heuristics.
Note that scheduling in Task \#2 is more challenging than in Task \#1 due to a greater variety and volume of target blocks. 
Furthermore, moving enemies and agent deaths from attacks in Task \#3 make planning non-trivial, even for the oracle.

Overall, we observe the following trends.
First, most tasks fail in async mode, as the average 20-second API delay frequently exceeds our benchmark's time-to-failure constraints (\cref{fig:2-metric_distributions}(d)). 
Second, in sync mode, \oursagent{}-\textbf{centralized} generally outperforms \textbf{distributed}. 
In distributed topology, inter-agent communication and inference overheads result in most of the time budget being spent for planning, leaving insufficient time for actions (detailed in \cref{fig:4-efficiency-scaling,fig:4-efficiency-breakdown}). 
Finally, even in sync mode, \textbf{centralized} falls short of the oracle's success rate. 
This highlights the difficulty of planning multi-agent collaboration in dynamic environments using partial observations rather than global ground-truth knowledge. 
These highlight the necessity of an efficient multi-agent coordination policy coupled with fast and accurate LLM planning.
Refer to Supp. \cref{sec:suppl-timeline-ex} for operational timeline examples.

\subsection{Ablation Study}

\begin{figure}[t]
  \centering
  \includegraphics[width=0.98\textwidth]{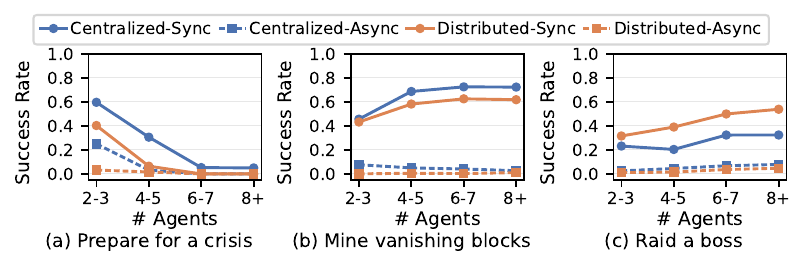}
  \vspace{-8pt}
  \caption{Task success rate across different numbers of agents.}
  \label{fig:4-agent-scaling}
  \vspace{-13pt}
\end{figure}

We analyze the factors behind \oursagent{}'s low success rate. Given consistent trends across both evaluated LLMs, we present results for GPT-5.1.

\noindent \textbf{Scaling with team size.}
\cref{fig:4-agent-scaling} shows the task success rate across different numbers of agents. 
For ``raid a boss'' and ``mine vanishing blocks'' tasks, increasing the number of agents improves enemy-killing or mining throughput, leading to higher success rates. 
In contrast, in ``prepare for a crisis'', increasing the number of agents increases both the required survival blocks to mine and the shelter size.
Since the success condition requires all agents to survive, the success rate tends to decrease.
Across all tasks, asynchronous mode generally yields lower success rates due to LLM inference latency (typically around 20 seconds per API call).

\begin{figure}[t]
  \centering
  \includegraphics[width=\textwidth]{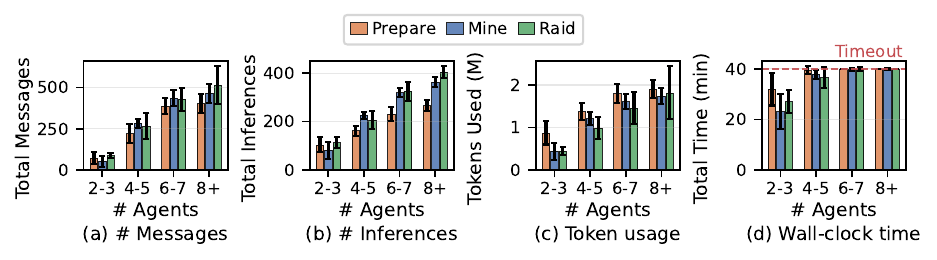}
  \vspace{-20pt}
  \caption{System costs of \textbf{\oursagent{}-distributed} in synchronous mode.}
  \label{fig:4-efficiency-scaling}
  \vspace{-12pt}
\end{figure}

\noindent \textbf{System costs.}
Figure~\ref{fig:4-efficiency-scaling} reports the communication and inference costs of the distributed baseline in synchronous mode.
Because agents can broadcast messages to multiple peers, message volume and LLM inference calls increase rapidly as team size increases (Figures~\ref{fig:4-efficiency-scaling}(a) and (b)).
Consequently, token usage and wall-clock time spike, often approaching the 40-minute simulation timeout (Figures~\ref{fig:4-efficiency-scaling}(c) and (d)).
This highlights the need for efficient communication protocols and group formation strategies in distributed multi-agent systems.

\begin{figure}[t]
  \centering
    \begin{subfigure}[t]{0.315\textwidth}
      \centering
      \includegraphics[width=\textwidth]{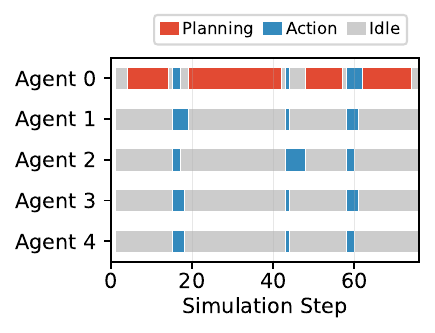}
      \vspace{-15pt}
      \caption{\textbf{\oursagent{}-centralized} timeline example.}
      \label{fig:efficiency-gantt}
    \end{subfigure}
    \hspace{1pt}
    \hfill
    \begin{subfigure}[t]{0.32\textwidth}
      \centering
      \includegraphics[width=\textwidth]{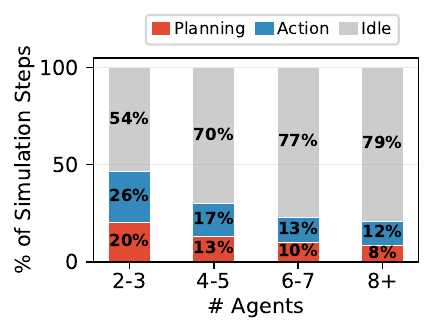}
      \vspace{-15pt}
      \caption{\textbf{\oursagent{}-centralized} step breakdown.}
      \label{fig:breakdown-centralized}
    \end{subfigure}
    \hspace{1pt}
    \hfill
    \begin{subfigure}[t]{0.32\textwidth}
      \centering
      \includegraphics[width=\textwidth]{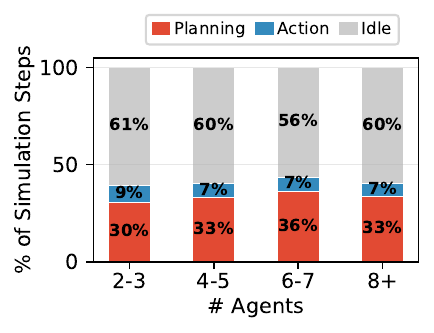}
      \vspace{-15pt}
      \caption{\textbf{\oursagent{}-distributed} step breakdown.}
      \label{fig:breakdown-distributed}
    \end{subfigure}
    \vspace{-2pt}
    \caption{Step-level planning, action, and idle-time breakdown in asynchronous mode.}
    \label{fig:4-efficiency-breakdown}
    \vspace{-10pt}
\end{figure}

\noindent \textbf{Coordination overhead.}
Figure~\ref{fig:4-efficiency-breakdown} breaks down asynchronous mode execution into planning, action, and idle steps.
In the centralized baseline, agents remain idle while the central planner waits for LLM inference (typically around 20 seconds per API call); additional idle time arises as agents' assigned actions vary in duration (Figure~\ref{fig:4-efficiency-breakdown}(a)).
This synchronization bottleneck becomes more pronounced as team size grows, leading to a larger idle fraction (Figure~\ref{fig:4-efficiency-breakdown}(b)).
The distributed baseline avoids a single central planner, but spends more time in planning because agents must communicate and negotiate before acting (Figure~\ref{fig:4-efficiency-breakdown}(c)).
Overall, these results highlight the need for agents to form appropriate coordination topologies and optimize both planning and action execution times.
\section{Limitations and Future Work}
\label{sec:5-discussion}

\noindent \textbf{Observation modality}.
We use distance-limited structured semantic sensors to focus on the collaborative planning capabilities of LLM agents. 
We plan to extend \oursbenchmark{} to multimodal inputs (e.g., first-person-view video and audio) to evaluate VLM agents.

\noindent \textbf{Agent-environment dynamics}.
\ours{} supports complex agent-conditioned interactions (e.g., inventory-based movement slowdowns). 
We will leverage these primitives to test collaboration under complex, cascading environmental changes driven by agent actions.

\section{Related Work}
\label{sec:6-related-work}

\noindent \textbf{Minecraft agents.}
While numerous works leverage Minecraft to develop autonomous agents~\cite{wang2023voyager, bolton2025sima, magne2026nitrogen, li2025jarvis, cai2024minestudio, gong2024mindagent, fan2022minedojo}, most focus on single-agent scenarios. 
Recent multi-agent extensions~\cite{long2024teamcraft, yu2024mineland, white2025minecollab, dong2024villageragent, schipper2025pillagerbench} mostly feature static environments, homogeneous agents, and tasks where collaboration is optional or lacks real-time constraints. 
Bridging this gap from real-world collaboration, \oursbenchmark{} introduces time-sensitive, complementary tasks requiring strictly mandatory coordination.

\noindent \textbf{Multi-agent collaboration.}
Broader evaluations of multi-agent collaboration in domains like mathematics, science, or coding~\cite{zhuge2024gptswarm, chen2024agentverse, yu2024researchtown} often lack environmental dynamics and explicit time limits. 
\oursbenchmark{} strictly demands both decision accuracy and low latency. 
Other orthogonal efforts improving multi-agent efficiency—such as task plan search~\cite{zu2025collaborative,zhang2025towards}, communication topologies~\cite{zhu2025multiagentbench, li2025exponential, qian2024scaling, zhuge2024gptswarm}, and resource-aware planning~\cite{yang2025bamas, cai2025agentbalance}—can be seamlessly integrated to tackle the real-time challenges in our benchmark.

\section{Conclusion}
\label{sec:7-conclusion}

We presented \oursbenchmark{}, a Minecraft benchmark evaluating \emph{time-sensitive complementary collaboration}. 
To capture the core characteristics of real-world collaboration (agent heterogeneity, mandatory coordination, dynamic environments, and real-time constraints), we developed an extensible orchestration framework and an automated task generation pipeline. 
Baseline evaluations reveal significant performance gaps in current LLMs, highlighting the critical need for more accurate and latency-efficient multi-agent systems.

\textbf{Acknowledgement.} 
\rev{This study is conducted for research only, not for an actual Minecraft product.}

\newpage
\bibliographystyle{abbrv}
\bibliography{main}


\newpage
\appendix
\appendix
\setcounter{page}{1}

\section{Task Metadata Schema}
\label{sec:suppl-metadata-schema}

\cref{list:2-metadata-full-example} shows the full task metadata for ``prepare for a crisis'' task.
It is mainly composed of four sections:
(i) \emph{task description} to specify the task goal and guidance, 
(ii) \emph{environment} to specify the simulation world layout (e.g., material, regions with specific interaction effects),
(iii) \emph{events} to specify the runtime dynamic event's trigger, type, and pattern, and
(iv) \emph{agents} to specify the attributes of each agent.
Overall, the interface is designed to be declarative and intuitive, making it easy for users to define custom tasks and for LLMs to generate valid outputs.

\begin{lstlisting}[caption={Example metadata for ``prepare for a crisis'' task.\vspace{0pt}},  frame=single, breaklines=true,  label={list:2-metadata-full-example},
aboveskip=0pt, belowskip=0pt] 
task:
  type: prepare_crisis
  goal: "Identify the type and origin of the crisis and build a shelter to survive"
  guidance: # can add
    text: |
      Each agent has different tools, speeds, and perception ranges.
    
events:
  - id: lava_wave
    trigger:
      start: 30
      end: 200
    actions:
      - type: progressive_fill
        block: lava
        area:
          min: {x: -30, y: 64, z: 0}
          max: {x: 40, y: 64, z: 40}
        direction: east   # starts at x=-30, moves toward x=40
        speed_bps: 1          # 1 blocks per trigger

agents:
  count: 4
  spawn:
    - name: MineflayerBot0
      position: [8, 64, 18]
      inventory:
        diamond_pickaxe:
          count: 1
          unbreakable: true
      capabilities:
        max_health: 40
        speed_bps: 4.3           # normal speed (~4.317 default)
        perception_range: 24     # reduced for partial observability
        effects:
          - fire_resistance
    - name: MineflayerBot1
      position: [12, 64, 18]
      inventory:
        diamond_axe:
          count: 1
          unbreakable: true
      capabilities:
        max_health: 30
        speed_bps: 6.5           # fast (~1.5x normal)
        perception_range: 16     # narrow scan
        effects:
          - fire_resistance
    - name: MineflayerBot2
      position: [8, 64, 22]
      inventory:
        iron_pickaxe:
          count: 1
          unbreakable: true
      capabilities:
        speed_bps: 3.0           # slow (~0.7x normal)
        perception_range: 20     # limited scan
    - name: MineflayerBot3
      position: [12, 64, 22]
      inventory:
        golden_axe:
          count: 1
          unbreakable: true
      capabilities:
        speed_bps: 8.0           # very fast (~1.85x normal)
        perception_range: 10     # very narrow scan - must get close

environment:
  max_steps: 100
  
  world:
    world_type: flatgrass
    difficulty: hard
    time: day
    gamerules:
      doMobSpawning: false
      doImmediateRespawn: false
      doDaylightCycle: false

  chest:
    position: [0, 64, 0]
  
  entities:
    hostiles: []
  
  materials:
    grid:
      # ===== FIRE-RESISTANT BLOCKS (scattered far from spawn) =====
      - block: cobblestone
        position: [20, 64, 30]
        width: 4
        height: 1
        depth: 5
      - block: cobblestone
        position: [25, 64, 35]
        width: 3
        height: 1
        depth: 4
      - block: cobblestone
        position: [10, 64, 40]
        width: 4
        height: 1
        depth: 5
      - block: stone
        position: [5, 64, 30]
        width: 3
        height: 1
        depth: 3
      - block: stone
        position: [10, 64, 25]
        width: 4
        height: 1
        depth: 3
      - block: stone_bricks
        position: [0, 64, 20]
        width: 3
        height: 1
        depth: 4
      - block: stone_bricks
        position: [25, 64, 5]
        width: 3
        height: 1
        depth: 3
      - block: bricks
        position: [20, 64, -12]
        width: 3
        height: 1
        depth: 3
      
      # ===== FLAMMABLE BLOCKS (TRAP - will burn!) =====
      - block: oak_log
        position: [-10, 64, 10]
        width: 3
        height: 1
        depth: 4
      - block: oak_log
        position: [15, 64, 20]
        width: 3
        height: 1
        depth: 4
      - block: oak_planks
        position: [15, 64, 5]
        width: 4
        height: 1
        depth: 3
      - block: oak_planks
        position: [-5, 64, 10]
        width: 3
        height: 1
        depth: 4
\end{lstlisting}

\section{User Specification Example for Automated Benchmark Generation}
\label{sec:suppl-user-specification}

\cref{list:2-user-specification-example} shows an example YAML interface for user specification used to generate task metadata with LLMs. The configuration file is parsed and converted into a text prompt for the LLM.
Users can specify the following components:
(i) \emph{task description} to define the task type,
(ii) \emph{variable parameters} to define the parameter space and their possible values (parameter names must match those in the task metadata),
(iii) \emph{constraints} to enforce formatting requirements for the generated metadata,
(iv) \emph{generation instructions} to guide the LLM to reflect the properties of our proposed time-sensitive complementary collaboration tasks,
(v) \emph{diversity} to encourage balanced sampling across varying parameters, and
(vi) \emph{references} to provide supporting documents (e.g., metadata schema, Minecraft physics rules, and interaction mechanisms).

\begin{lstlisting}[caption={Example user specification for ``prepare for a crisis'' task.\vspace{0pt}},  frame=single, breaklines=true,  label={list:2-user-specification-example},
aboveskip=0pt, belowskip=0pt] 
task_type: prepare_crisis
description: "Multi-agent team prepares for an incoming crisis by gathering resources and building defenses"

# Parameter variation space in the task metadata
variable_parameters:
  # Crisis Event Configuration
  events.actions.type:
    description: "Crisis event type"
    allowed: [progressive_fill]

  events.actions.block:
    description: "Block type for progressive_fill events"
    allowed: [lava, water, powder_snow]

  events.actions.direction:
    description: "Fill direction (horizontal only)"
    allowed: [east, west, north, south]

  events.actions.area.min.y:
    description: "Area min Y coordinate (must be 64)"
    range: [64, 64]

  events.actions.area.max.y:
    description: "Area max Y coordinate (must be 64)"
    range: [64, 64]

  events.actions.speed:
    description: "Blocks filled per trigger"
    range: [1, 5]

  events.trigger.start:
    description: "Step when crisis begins"
    range: [3, 20]

  events.trigger.end:
    description: "Step when crisis ends"
    range: [20, 200]

  # Agent Configuration
  agents.count:
    description: "Team size"
    range: [1, 8]

  agents.spawn.inventory:
    description: "Agent starting tools and armor"
    allowed: [wooden_pickaxe, stone_pickaxe, iron_pickaxe, diamond_pickaxe, golden_pickaxe, netherite_pickaxe, wooden_axe, stone_axe, iron_axe, diamond_axe, netherite_axe, leather_boots]
    match: keys

  agents.spawn.capabilities.max_health:
    description: "Agent HP"
    range: [10, 60]

  agents.spawn.capabilities.speed_bps:
    description: "Agent movement speed multiplier"
    range: [0.5, 2.0]

  agents.spawn.capabilities.effects.type:
    description: "Potion effect types"
    allowed: [fire_resistance]

  # Environment Configuration
  environment.materials.grid.block:
    description: "Building blocks scattered in world"
    allowed: [stone, cobblestone, stone_bricks, bricks, deepslate, iron_block, gold_block, diamond_block, obsidian, crying_obsidian, netherite_block, oak_log, birch_log, spruce_log, dark_oak_log, oak_planks]

  environment.materials.grid.height:
    description: "Height of material piles"
    range: [1, 1]

  # Environment
  environment.max_steps:
    description: "Time limit for the scenario"
    range: [50, 200]

# Formatting constraints when generating the metadata
constraints: |
  CRITICAL YAML FORMATTING RULES

  1. Coordinates must be inline lists: position: [8, 64, 18]
  2. Area centers must be inline lists: center: [0, 64, 20]
  3. Area min/max must be inline lists
  4. Keep simple key-value pairs on one line
  5. Use quoted strings for descriptions
  6. Keep the same indentation and structure as the template
  7. The events list must contain exactly one event with one progressive_fill action
  8. Crisis area must use Y=64 for both min and max
  9. Direction must be east, west, north, or south

# Instructions to guide the LLM to reflect the characteristics of time-sensitive, complementary collaboration tasks
generation_instructions: |
  Generate diverse crisis preparation scenarios by varying the parameters listed in variable_parameters.

  Crisis type distribution:
  Use lava, water, or powder_snow. If not assigned, choose randomly.

  Agent count diversity:
  Vary agent count across the full range (2-8).

  Design guidelines:

  Agent heterogeneity:
  - Each agent must have different attributes
  - Mix tool types and movement speeds
  - For lava scenarios, give some agents fire_resistance

  Collaboration necessity:
  - N = (blocks_to_gather + blocks_to_place) / best_agent_throughput / T_max
  - N must be greater than 1.0 so a single agent cannot complete alone

  Environment dynamicity:
  - Increase fill speed or trigger frequency to increase dynamics

  Time-to-failure:
  - Lower distance or higher speed increases urgency

# Enforces dataset diversity
diversity_round_robin:
  - field: events.actions.block
    label: Crisis Type Assignment
    instruction: "Adapt all fields to match the assigned crisis type."

  - field: agents.count
    label: Agent Count Assignment
    values: [2, 2, 2, 3, 3, 3, 4, 4, 5, 7, 8]
    instruction: "Use exactly the assigned number of agents."

# Reference documents (task metadata schema, minecraft's physics rules and interaction mechanisms, etc.)
references:
  - metadata_schema.md
  - mining_items.md
  - interaction_effects.md
\end{lstlisting}

\section{Verification Criteria for \oursbenchmark{} Generation}
\label{sec:supp-verification}

In this section, we provide a detailed explanation of the verification criteria for the automated benchmark generation pipeline shown in Tab. 3 of our submission.

\subsection{Task \#1: Prepare for a Crisis}

For this task, we first verify if at least one agent has necessary item to mine the target survival block, by checking the following:

\begin{equation*}
    \resizebox{0.45\linewidth}{!}{$
        \max_{n}(\tau_{E_n}) \ge \tau_{req}(b), \quad \forall b \in \mathcal{B}_{target},
    $}
\end{equation*}

\noindent where $\tau_{E_n}$ is the mining tier of the tool of the $n$-th agent, 
$\tau_{req}(b)$ denotes the required mining tier for block type $b$, and $\mathcal{B}_{target}$ is the set of target survival block types. 
This condition ensures that at least one agent possesses a tool capable of mining the required survival blocks.

Next, we verify whether a sufficient number of survival blocks are placed in the map. Specifically, we check the following:

\begin{equation*}
    \resizebox{0.35\linewidth}{!}{$
        N_{required} \geq (H+1) \cdot N_{agent},
    $}
\end{equation*}

\noindent where $N_{required}$ denotes the number of survival blocks available in the environment (e.g., stone blocks when the crisis type is lava flood), $N_{agent}$ is the number of agents, and $H$ is the height of the crisis.
Since agents must build a shelter higher than the crisis height and stand on top of it to survive, this condition verifies that a sufficient number of blocks are available to construct such a structure.

Finally, we verify whether the agents have sufficient time to build a shelter before the crisis hits by evaluating the following condition:

\begin{equation*}
    \resizebox{0.4\linewidth}{!}{$
        T_{crisis} \geq \alpha \cdot (T_{move}+T_{mine}+T_{construct}).
    $}
\end{equation*}

\noindent 
Here, $T_{crisis}$ is the time it takes for the crisis to reach the build site, computed as $T_{crisis} = t_0 + {L_{area}}/{v_{fill}}$. 
$t_0$ is the crisis start time, $L_{area}$ is the length of the crisis area along the spread direction, and $v_{fill}$ is the fill speed in blocks per step. 
We assume the build site is located at the farthest end of the crisis area (i.e., the last point the crisis reaches).

The right-hand side of the inequality decomposes the total required preparation time into two phases:

\begin{itemize}[leftmargin=*, topsep=0pt, itemsep=0pt, parsep=0pt]
\item $T_{move}+T_{mine} = \max_{a} W_a$ is the makespan of the material gathering phase (moving to the blocks and mining them). 
Each agent $a$ is greedily assigned a subset of material block piles to visit. 
Piles are sorted by proximity to the build site and assigned to a compatible agent (i.e., one holding a tool of sufficient tier) that currently has the lowest accumulated workload. 
Agent $a$'s workload is defined as:$$W_a = \sum_{i \in \mathcal{P}_a} \left( \frac{d(p^a_{i-1}, p^a_i)}{v_a} + n^a_i \cdot 1.5 \cdot \frac{h_{b_i}}{s^a_{b_i}} \right) + \frac{d(p^a_{|\mathcal{P}_a|}, \mathbf{s})}{v_a}$$where $\mathcal{P}_a$ is the ordered set of piles assigned to agent $a$, $p^a_i$ is the position of the $i$-th pile ($p^a_0$ being the agent's spawn position), $v_a$ is the agent's movement speed, $n^a_i$ is the number of blocks mined from that pile, $h_{b_i}$ is the block hardness, $s^a_{b_i}$ is the mining speed of agent $a$'s best tool for block type $b_i$, and $\mathbf{s}$ is the build site position. 
The bottleneck agent determines the overall gather time.

\item $T_{construct} = \frac{0.5 \cdot N_{required}}{N'_{agents}}$ estimates the parallel construction time, assuming $0.5$ seconds per block placement (the default value in Minecraft). 
Here, $N_{required}$ denotes the total number of blocks required for the shelter floor and access stairs, and $N'_{agents}$ is the number of active agents equipped with a valid mining tool (and thus assumed to participate in the construction).

\end{itemize}
Since the above terms are estimates, we introduce $\alpha \geq 1$ as a time safety margin. In the context of task generation, a larger $\alpha$ enforces a longer delay before the crisis arrives, thereby creating an easier task for the agents.

\subsection{Task \#2: Mine Vanishing Blocks}

For this task, we first verify if at least one agent has necessary item to mine the target survival block, by checking the following:

\begin{equation*}
    \resizebox{0.45\linewidth}{!}{$
        \max_{n}(\tau_{E_n}) \ge \tau_{req}(b), \quad \forall b \in \mathcal{B}_{target},
    $}
\end{equation*}

\noindent where the variable notations are the same as in the ``Prepare for a crisis'' task.

Next, we verify whether the block lifetimes are sufficiently long for agents to reach and mine them by evaluating the following condition:

\begin{equation*}
    \resizebox{0.45\linewidth}{!}{$
        \beta_1 \cdot (\min_{n} (T_{move}+T_{mine}^{n,i})) \leq T_{life}^i, \; \forall i,
    $}
\end{equation*}

\noindent Here, for the $i$-th target block, $T_{move}$ is the estimated travel time from the mean position of the mining-capable agents to the spawn area. 
This is calculated using the distance to the center of the target block spawn area plus $\frac{2}{3}R$, which represents the expected offset within a spawn radius $R$.
Furthermore, $T_{mine}^{n,i}$ is the time required for agent $n$ to mine one block of type $i$, and $T_{life}^i$ is the lifetime of block type $i$ before despawning. 
The minimization is computed over all agents $n$ capable of mining block type $i$.
Since the actual positions of the agents and blocks are dynamic at runtime, we introduce a safety margin $\beta_1 \geq 1$. 
Enforcing a larger $\beta_1$ during task generation requires longer block lifetimes, thereby making the task easier for the agents.

Finally, we verify whether the total number of spawned target blocks is sufficient to meet the target count by checking the following condition:

\begin{equation*}
\resizebox{0.22\linewidth}{!}{$
\textstyle N_{spawn}^i \geq \beta_2 \cdot N_{goal}^i \; \forall i,
$}
\end{equation*}

\noindent where $N_{spawn}^i$ is the total number of spawned blocks of type $i$, and $N_{goal}^i$ is the required number of blocks of type $i$ as specified by the task goal. 
We introduce $\beta_2 \geq 1$ as a supply safety margin to account for blocks that may despawn before the agents can reach and mine them.

\subsection{Task \#3: Raid a Boss}

For this task, we evaluate whether the agents have sufficient attack damage to defeat the boss and its spawned minions within the simulation duration.
Specifically, we verify the following conditions:

\begin{equation*}
    \resizebox{0.5\linewidth}{!}{$
        \gamma \cdot T_{max} \cdot \sum_{n} D_n \geq HP_{B} + \sum_i ( N_i\cdot HP_M^i),
    $}
\end{equation*}

\noindent where $T_{max}$ denotes the total simulation duration, $D_n$ is the attack damage of the $n$-th agent, $HP_B$ and $HP_M^i$ are the life points of the boss and the $i$-th minion, respectively, and $N_i$ is the total number of generated $i$-th minion type.
The left-hand side of the equation represents the maximum damage that the agents can theoretically inflict. 
In practice, however, both agents and enemies continuously move, making it impossible to achieve this maximum damage.
To account for this, we introduce an efficiency margin $\gamma$ as a soft constraint. 
During the verification stage, the desired range of $\gamma$ can be specified to control the task difficulty, where a larger $\gamma$ indicates an easier task.

\section{Feasibility Margin Sensitivity}
\label{sec:suppl-margin_sensitivity}

\begin{figure}[t]
    \centering
    \includegraphics[width=0.65\columnwidth]{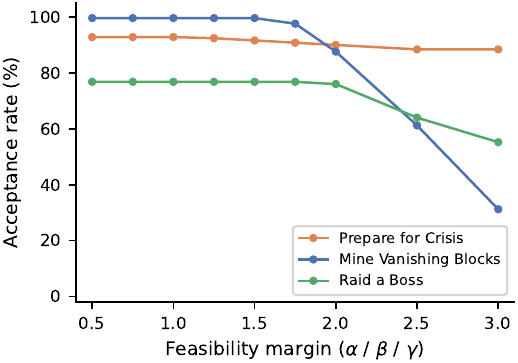}
    \caption{Acceptance rate vs.\ feasibility margin $\alpha$ for each task category. Higher margins impose stricter feasibility constraints, reducing the fraction of retained configurations.}
    \label{fig:margin_sensitivity}
\end{figure}

Figure~\ref{fig:margin_sensitivity} shows how the acceptance rate of generated configurations varies as the feasibility margin increases. A higher margin imposes stricter constraints—requiring agents to complete objectives with a larger safety buffer—thereby filtering out configurations that are only marginally feasible. At our chosen threshold of $\alpha{=}2.0\times$, acceptance rates remain above 87\% across all tasks, and degrade gracefully as $\alpha$ increases further. Notably, \textit{Prepare for Crisis} does not reach 100\% acceptance even at the lowest margin ($\alpha{=}0.5$), plateauing around 94\%. This is because a small fraction of LLM-generated configurations assign crisis types (e.g., lava, water, powder snow) for which no agent possesses the appropriate tool to harvest the corresponding survival block, making the scenario structurally infeasible regardless of margin.

\section{\oursbenchmark{} Generated Dataset Examples}
\label{sec:suppl-dataset-ex}

\begin{figure}[t]
\begin{center}
\centerline{\includegraphics[width=1\columnwidth]{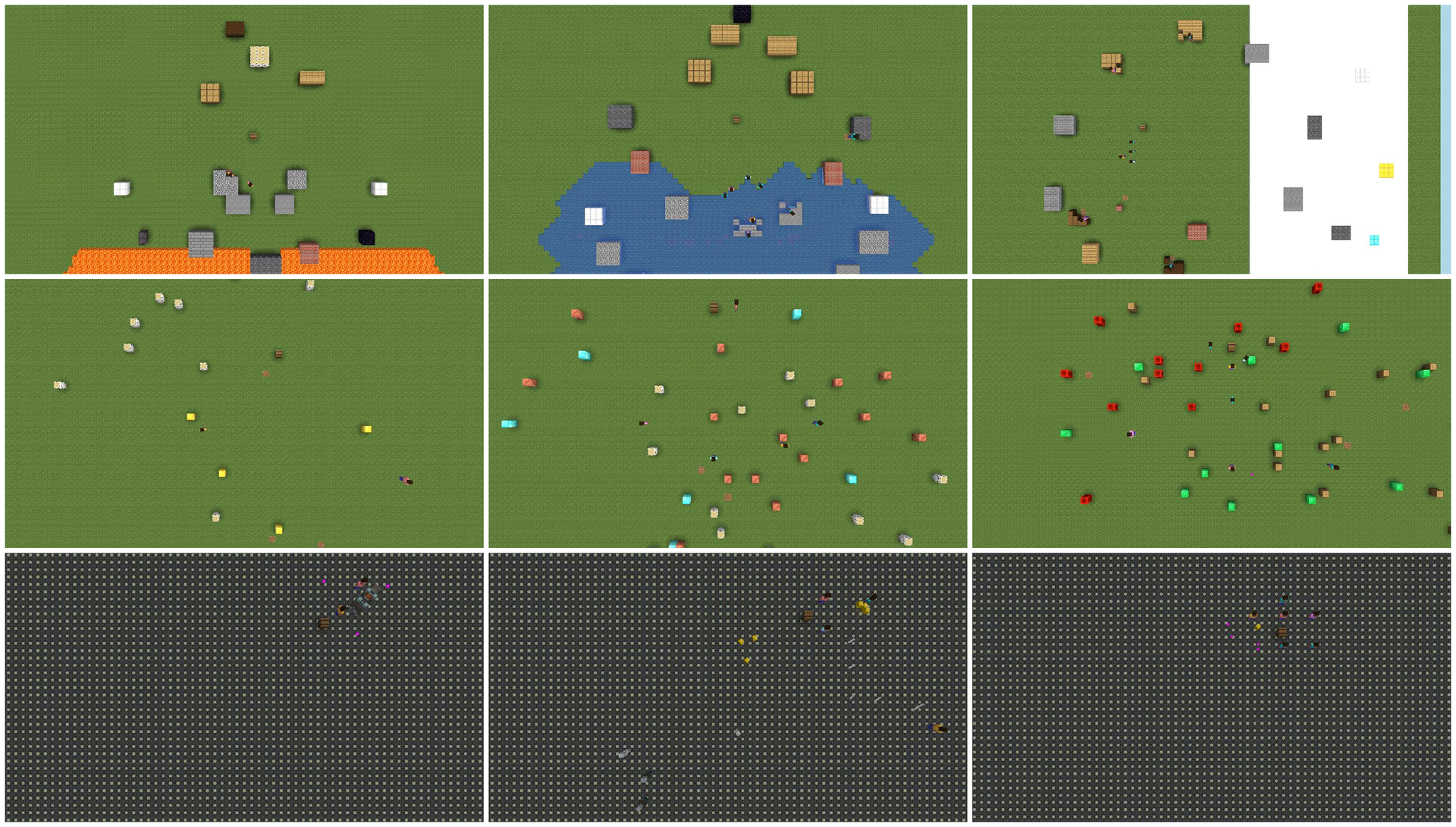}}
\caption{Examples in \oursbenchmark{}. Top row: ``prepare for a crisis'' task, middle row: ``mine vanishing blocks'' task, bottom row: ``raid a boss'' task.}
\label{fig:suppl-dataset-examples}
\end{center}
\vskip -0.2in
\end{figure}

\cref{fig:suppl-dataset-examples} shows example tasks from \oursbenchmark{}. 
Overall, the results demonstrate that the LLM generates diverse task variations.
For the ``prepare for a crisis task'', multiple crisis types with different effects are generated (e.g., lava flood which burns wooden blocks, water flood which spreads rapidly and can sweep agents away, snow avalanche that slows down agents). 
The intensity and frequency of the damage received when agents come into contact with the crisis are controlled by the \ours{} framework's dynamic environment manager.
In addition, the origin and propagation speed of the crisis, material types, quantities, and locations of the shelter blocks, as well as the agent configurations, vary across tasks, resulting in different difficulty levels.
For the ``mine vanishing blocks task'', the target block types and target quantities vary, along with the types and numbers of blocks placed in the environment and the agent configurations.
Finally, for the ``raid a boss'' task, the monster spawn patterns (e.g., a few strong monsters or many weaker ones) and the agent configurations differ across tasks.
Since generating diverse task configurations requires jointly considering multiple parameters and their interactions, manually designing such variations is challenging. Our LLM-based automated benchmark generation approach therefore provides an efficient solution.

\section{\oursagent{} Action Spaces}
\label{sec:suppl-agent-action}

\begin{table}[t]
\centering 
\caption{Agent toolset for each task.}
\label{tb:2-toolset}
\begin{small}
\setlength{\tabcolsep}{6pt}
\resizebox{\linewidth}{!}{%
\begin{tabular}{llll}
\toprule
{\bf Task} & {\bf Tool name} & {\bf Parameters} & {\bf Description} \\
\midrule
\multirow{6}{*}{\makecell[l]{Prepare\\for a crisis}}
  & \emph{scout\_blocks\_at}  & \textit{target\_pos}, \textit{max\_distance}              & Move to location and search for blocks \\
  & \emph{mine\_blocks\_at}   & \textit{block\_positions}                                 & Mine blocks at specific positions \\
  & \emph{build\_floor}       & \textit{center\_pos, width, depth, height}                & Build a layered floor platform \\
  & \emph{move\_to}           & \textit{target\_pos}                                      & Move to a location \\
  & \emph{deposit\_to\_chest} & \textit{chest\_pos, items, quantities}                    & Deposit items into shared chest \\
  & \emph{get\_from\_chest}   & \textit{chest\_pos, items, quantities}                    & Get items from shared chest \\
\midrule
\multirow{4}{*}{\makecell[l]{Mine vanishing\\ blocks}}
  & \emph{scout\_blocks\_at}  & \textit{target\_pos, max\_distance}                       & Move to location and search for blocks \\
  & \emph{mine\_blocks\_at}   & \textit{block\_positions}                                 & Mine blocks at specific positions \\
  & \emph{deposit\_to\_chest} & \textit{chest\_pos, items, quantities}                    & Deposit items into shared chest \\
\midrule
\multirow{8}{*}{Raid a boss}
  & \emph{move\_to}           & \textit{target\_pos}                                      & Move to a location \\
  & \emph{find\_entities}     & \textit{entity\_types, count, max\_distance}              & Scan for nearby entities \\
  & \emph{attack}             & \textit{entity\_type}                                     & Attack a target entity \\
  & \emph{equip\_item}        & \textit{item}                                             & Equip a weapon \\
  & \emph{use\_item}          & \textit{item}                                             & Use a consumable item \\
  & \emph{deposit\_to\_chest} & \textit{chest\_pos, items, quantities}                    & Deposit items into shared chest \\
  & \emph{get\_from\_chest}   & \textit{chest\_pos, items, quantities}                    & Get items from shared chest \\
\bottomrule
\end{tabular}
}
\end{small}
\vskip -0.2in
\end{table}

\cref{tb:2-toolset} summarizes the agent toolset for each task.
Each tool is implemented as a JavaScript function, which is generated by the LLM and executed by Mineflayer.
For the ``prepare for a crisis task'', agents can move and scout nearby blocks (including both crisis types and candidate survival blocks) using the \emph{scout\_blocks\_at()} function, where \emph{max\_distance} specifies the agent’s maximum perception range. 
Agents can mine target survival blocks using \emph{mine\_blocks\_at()}, and construct a shelter using \emph{build\_floor()}. 
Agents that do not participate in the mining process can retreat to the shelter via \emph{move\_to()}.
Item exchange is performed through a chest, whose location is predefined in the task metadata, using the \emph{deposit\_from\_chest()} and \emph{get\_from\_chest()} functions.

For the ``mine vanishing blocks'' task, the action set is simpler, consisting of three functions for scouting and mining blocks, and depositing mined blocks into the chest. The task is considered successful when the chest contains the target block type in the required quantity.

For the ``raid a boss task'', agents similarly scan nearby enemies, equip or use appropriate items, and exchange items through the chest.
For the \emph{attack()} function, the input is not a fixed position because monsters continuously move. 
Instead, the function searches for the nearest target entity within the vicinity and performs tracking and attacking based on the object reference of that entity. 
The attack process terminates when either a timeout (15 seconds) is reached or the target entity is defeated. 
This functionality is implemented using Mineflayer-pvp~\cite{mineflayer-pvp} module.

\section{Agent System Prompts}
\label{sec:suppl-system-prompts}

We provide the full system prompt templates for both baseline agent architectures evaluated in our experiments.
Template variables in \texttt{\{curly\_braces\}} (e.g., \texttt{\{agent\_name\}}, \texttt{\{task\_type\}}) are populated at runtime with the corresponding agent state, task specification, and environment context.
\cref{list:prompt-centralized} shows the centralized baseline prompt, where a single LLM call receives all agents' states and produces a joint plan.
\cref{list:prompt-distributed} shows the distributed baseline prompt, where each agent independently makes decisions and coordinates through an explicit negotiation protocol.

\subsection{Centralized Baseline}

\begin{lstlisting}[caption={System prompt template for the centralized baseline.\vspace{0pt}},
  label={list:prompt-centralized}, frame=single, breaklines=true,
  basicstyle=\footnotesize\ttfamily, escapeinside={(*@}{@*)},
  aboveskip=0pt, belowskip=0pt]
You are a centralized controller managing ALL {num_agents} agents in a Minecraft team.

You receive every agent's state, capabilities, and action history. You produce a
coordinated plan for ALL agents simultaneously. Each agent will execute ONLY the
tasks you assign to it. Agents do NOT communicate with each other -- you are the
single decision-maker.

Your plans will be executed IMMEDIATELY. You will be called again when ALL agents
finish their assigned tasks.

## Strategy
Think holistically. You see the full picture -- every agent's position, health,
inventory, and capabilities. Use this to:
- Assign each agent to the sub-task that best fits its capabilities
- Avoid duplicate work (two agents mining the same block)
- Balance workload so agents finish at roughly the same time
- Anticipate coordination needs (e.g., one agent scouts while another builds)

## Output Format
Output a single JSON object:

{
  "reasoning": "Brief explanation of your overall strategy (2-3 sentences)",
  "agent_plans": {
    "{agent_name_0}": [
      {
        "id": "{agent_name_0}_task_1",
        "do": "action_name",
        "with": {},
        "after": [],
        "note": "Human-readable description"
      }
    ],
    (*@\textit{\color{gray}... (one entry per agent, same format)}@*)
  }
}

**Fields:**
- **reasoning**: Your overall coordination strategy (2-3 sentences).
- **agent_plans**: A dictionary mapping each agent name to its task list.
  - Each task list follows DAG format -- tasks execute in dependency order.
  - The "after" field must ONLY reference task IDs within the SAME agent's plan.
  - Task IDs must be prefixed with the agent's name.

CONSTRAINTS:
1. You MUST assign tasks to each alive agent.
2. Use ONLY positions from the provided knowledge -- NEVER invent coordinates.
3. If no blocks are known, assign find_blocks or scout_blocks_at first.
4. The "after" field must ONLY reference task IDs within the SAME agent's plan
   -- never cross-agent dependencies.
5. Output ONLY the JSON object, no additional text.
6. Every action requires its parameters. Missing parameters will cause FAILURE.
7. Do NOT assign tasks to DEAD agents. They are marked below.
8. BREAK REPETITION: If an agent's actions failed 2+ times, assign a
   different approach.
9. BALANCE WORKLOAD: Assign roughly equal work to each agent. You will not be
   called again until ALL agents complete their plans.

## Task Information
- Task Type: {task_type}
- Goal: {goal}

## Guidance
{guidance_text}

## Task Content
{task_content}

## Environment
{env_context}
{task_specific_context}
{available_actions}

- wait: Pause for a specified number of steps.
  - with: {duration: N} (1--50 steps)

## All Agents
{all_agents_context}

## Combined History (recent, all agents)
{combined_history}

## Current Step: {current_step} / {max_steps}
{dynamic_subgoals}
\end{lstlisting}

\subsection{Distributed Baseline}

\begin{lstlisting}[caption={System prompt template for the distributed baseline.\vspace{0pt}},
  label={list:prompt-distributed}, frame=single, breaklines=true,
  basicstyle=\footnotesize\ttfamily, escapeinside={(*@}{@*)},
  aboveskip=0pt, belowskip=0pt]
You are Agent {agent_name} (index {agent_idx}) in a fully distributed
{num_agents}-agent Minecraft team.

You decide your own actions autonomously -- you plan, communicate, negotiate,
and execute on your own judgment. By default there is no designated leader, but
you may PROPOSE or ACCEPT leadership, delegate tasks, or form any coordination
structure if the task benefits from it. Your plan will be executed IMMEDIATELY.

## Teamwork
You are part of a distributed team. You and your peers independently decide your
own actions, but effective teams NEGOTIATE before acting on shared decisions.
When a decision affects multiple agents -- role assignment, build location, target
priority, retreat -- propose your idea, wait for responses, then act on the outcome.

## Negotiation & Coordination
Use the propose -> wait -> read -> act pattern:

1. **Propose**: Send a [PROPOSAL] message describing your idea.
2. **Wait**: Use the `wait` action to pause while peers read and respond.
3. **Read**: After the wait, Conversation History will contain peer responses.
4. **Act**: Incorporate feedback -- adjust, accept, or counter-propose.

Negotiate when: assigning roles, choosing build locations, dividing territory,
forming strategy. Skip negotiation when: HP is critical, blocks are despawning,
or an attack is in progress -- act first and inform peers after.

### Responding to Proposals
When you see a [PROPOSAL] or [COUNTER-PROPOSAL] in Conversation History:
- **[AGREE]**: You accept. State what you will do under the agreed plan.
- **[DISAGREE]**: You reject. Explain why and suggest an alternative.
- **[COUNTER-PROPOSAL]**: You propose a modification. Explain the change.
Silence after a reasonable wait is treated as implicit agreement.

### Ongoing Coordination
Re-negotiate when: action fails 2+ times, your area runs dry, HP drops below
50%, task phase changes (e.g., scouting -> building), or a peer sends [HELP].
(*@\textit{\color{gray}... (detailed re-negotiation guidelines for each trigger)}@*)

### Topology
Your team can adopt any coordination structure -- you decide together:
- **Distributed** (default): Each agent decides independently.
- **Leader-worker**: One agent coordinates, others follow.
- **Tree / Hierarchy**: Sub-leaders for sub-teams.
Topology is not fixed. Any agent can propose a change at any time via [PROPOSAL].

## Communication
Messages are free-form natural language. Tag messages for clarity:
  [PROPOSAL]        -- role, plan, topology, or strategy suggestions
  [AGREE]           -- accepting a peer's proposal
  [DISAGREE]        -- rejecting a proposal (include reason + alternative)
  [COUNTER-PROPOSAL]-- proposing a modification to someone's plan
  [STATUS]          -- action results or progress updates
  (*@\textit{\color{gray}... (and [DISCOVERY], [HELP], [CONFLICT], [PLAN\_CHANGE])}@*)

### Communication Discipline
- No-repeat rule: Do NOT resend messages that already cover your situation.
- When sending a [PROPOSAL], follow it with a `wait` action to collect responses.
- When receiving a [PROPOSAL], respond with [AGREE], [DISAGREE], or
  [COUNTER-PROPOSAL]. Do not silently ignore it.
- Respect claims: If a peer claimed coordinates, pick different targets.
(*@\textit{\color{gray}... (additional collision-avoidance and target-claiming rules)}@*)

## Output Format
Output a single JSON object:

{
  "reasoning": "Brief explanation of your current decision",
  "my_role": "Your current role (e.g., 'Tank', 'Miner', 'Scout')",
  "plan": [
    {
      "id": "{agent_name}_task_1",
      "do": "action_name",
      "with": {},
      "after": [],
      "note": "Human-readable description"
    }
  ],
  "messages": [
    {"to": "MineflayerBot1", "content": "[PROPOSAL] I'll mine cobblestone
      at [5,64,10]. Can you mine stone at [20,64,8]?"},
    (*@\textit{\color{gray}... (one message per recipient)}@*)
  ]
}

**Fields:**
- **reasoning**: Why you chose this action (1-2 sentences).
- **my_role**: Your current team role. Use "Undecided" if no role yet.
- **plan**: Task sequence in DAG format. Can be empty [] if only messaging.
- **messages**: Messages to peers. Can be empty [].

CONSTRAINTS:
1. Plan ONLY tasks for YOURSELF -- do NOT include an "agent" field.
2. Use ONLY positions from your knowledge -- NEVER invent coordinates.
3. If no blocks are known, use find_blocks or scout_blocks_at first.
4. The "after" field must ONLY reference task IDs within YOUR OWN plan.
5. Output ONLY the JSON object, no additional text.
6. Every action requires its parameters. Missing parameters will cause FAILURE.
7. RESPECT CLAIMS: If a peer claimed coordinates, choose different targets.
8. BREAK REPETITION: If the same action fails 2+ times, try a different
   approach.

### Negotiation Example (propose -> wait -> act)

Agent proposes roles and waits for team input:
{
  "reasoning": "First decision. Coordinating roles before we start.",
  "my_role": "Undecided",
  "plan": [
    {"id": "{agent_name}_wait", "do": "wait",
     "with": {"duration": 5}, "after": [],
     "note": "Wait for team to respond to role proposal"}
  ],
  "messages": [
    {"to": "MineflayerBot1", "content": "[PROPOSAL] I have a netherite
      sword and high HP -- I'll tank. You have a bow -- ranged DPS?"},
    {"to": "MineflayerBot2", "content": "[PROPOSAL] I'll tank. You have
      golden apples -- healer/support?"}
  ]
}
(*@\textit{\color{gray}... (after wait, agent reads responses and executes the agreed plan)}@*)

## Task Information
- Task Type: {task_type}
- Goal: {goal}

## Guidance
{guidance_text}

## Task Content
{task_content}

## Environment
{env_context}
{task_specific_context}
{available_actions}

- wait: Pause for a specified number of steps. Use after sending a [PROPOSAL]
  to give peers time to respond.
  - with: {duration: N} (1--50 steps)

## Peer Agents
{peer_agents}

## History
{history}

## Your Current State
- Current Step: {current_step} / {max_steps}
{agent_state}

## Currently Executing Action
{current_action}

## Remaining Plan
{remaining_plan}
{dynamic_subgoals}
\end{lstlisting}


\section{\oursagent{} Operation Timeline Examples}
\label{sec:suppl-timeline-ex}

\begin{figure*}[ht]
    \centering
    \begin{subfigure}[b]{1\linewidth}
        \centering
        \includegraphics[width=\linewidth]{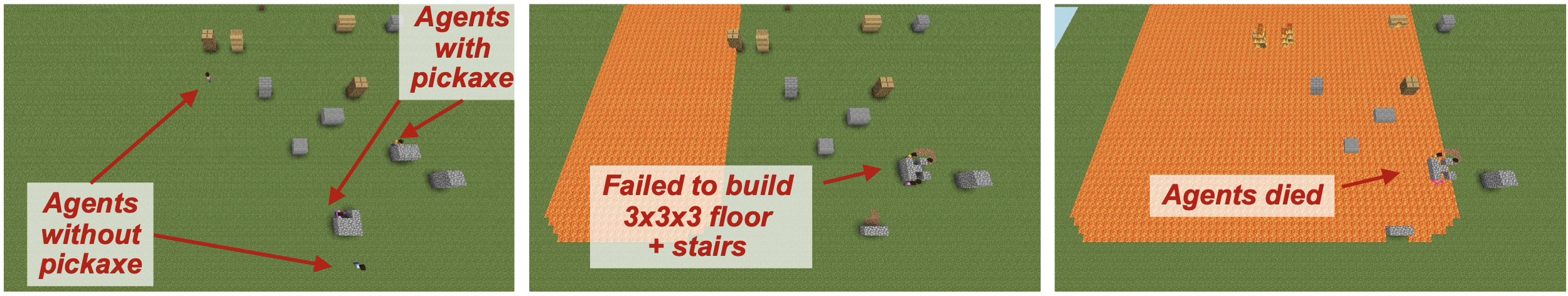}
        \caption{Prepare for a crisis.}
        \label{fig:4-timeline-ex-a}
    \end{subfigure}
    \hfill
    \begin{subfigure}[b]{1\linewidth}
        \centering
        \includegraphics[width=\linewidth]{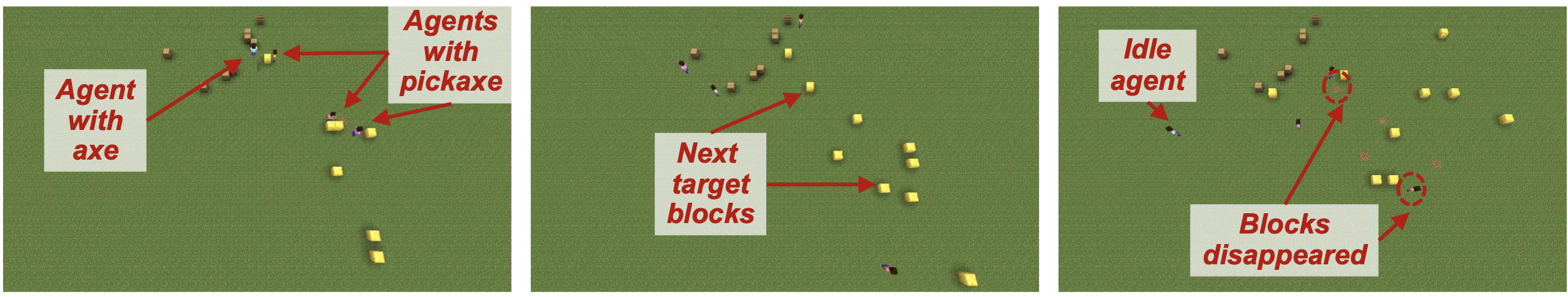}
        \caption{Mine vanishing blocks.}
        \label{fig:4-timeline-ex-b}
    \end{subfigure}
    \vskip -0.1in
    \caption{\oursagent{}-centralized operation timelines examples.}
    \label{fig:4-timeline-ex}
    \vskip -0.2in
\end{figure*}

\cref{fig:4-timeline-ex} shows example operational timelines of \oursagent{}-\textbf{centralized}.
\cref{fig:4-timeline-ex-a} shows an example of the ``prepare for a crisis'' task, where four agents need to build a shelter to survive from a lava flood. 
The central agent correctly selects stone blocks (instead of wood which burns) and assigns two agents with pickaxes to mine them, while sending the other two to scout the flood.
It also chooses a shelter location far from the crisis.
However, it instructs only one agent to build the shelter, leaving the other mining agent idle. 
As a result, the shelter remains incomplete due to insufficient materials. 

\cref{fig:4-timeline-ex-b} shows another example from the ``mine vanishing blocks'' task, where five agents need to mine 2 oak logs and 15 gold blocks. 
The central agent correctly assigns agents with axes and pickaxes to the corresponding targets.
However, it does not correctly account for the lifetime of the blocks when scheduling the mining actions; as a result, the blocks often disappear by the time an agent reaches them.
Meanwhile, while the two agents repeatedly fail to mine the disappeared blocks, the remaining agents stay idle; 
a more effective strategy would have been to proactively search for newly spawned blocks at other locations.

\end{document}